\documentclass[10pt,twocolumn,letterpaper]{article}

\usepackage{cvpr}              

\usepackage{lipsum}
\usepackage{multirow}
\usepackage{tabularx}
\usepackage{makecell}
\usepackage{amsmath}
\usepackage{color, colortbl}
\usepackage{nccmath}
\usepackage{float}
\usepackage{amssymb}
\usepackage{afterpage}
\usepackage[symbol]{footmisc}
\usepackage{tablefootnote}
\usepackage[accsupp]{axessibility} 
\usepackage[normalem]{ulem}

\usepackage[dvipsnames]{xcolor}

\definecolor{cvprblue}{rgb}{0.21,0.49,0.74}
\usepackage[pagebackref,breaklinks,colorlinks,citecolor=cvprblue]{hyperref}
\definecolor{tablegray}{rgb}{0.9,0.9,0.9}

\def\paperID{10040} 
\def\confName{CVPR}
\def\confYear{2025}

\title{Multi-modal Knowledge Distillation-based Human Trajectory Forecasting}

\author{Jaewoo Jeong$^1$, Seohee Lee$^1$, Daehee Park$^2$\textsuperscript{\dag}, Giwon Lee$^1$, and Kuk-Jin Yoon$^1$\\
$^1$Visual Intelligence Lab., KAIST, Korea\\
$^2$Intelligent Systems and Learning Lab., DGIST, Korea\\
}

\begin{document}
\maketitle
\renewcommand{\thefootnote}{\dag} 
\footnotetext{This work was completed as a Ph.D candidate at KAIST.}
\renewcommand{\thefootnote}{\arabic{footnote}} 
\begin{abstract}
Pedestrian trajectory forecasting is crucial in various applications such as autonomous driving and mobile robot navigation. 
In such applications, camera-based perception enables the extraction of additional modalities (human pose, text) to enhance prediction accuracy.
Indeed, we find that textual descriptions play a crucial role in integrating additional modalities into a unified understanding.
However, online extraction of text requires the use of VLM, which may not be feasible for resource-constrained systems.
To address this challenge, we propose a multi-modal knowledge distillation framework: a student model with limited modality is distilled from a teacher model trained with full range of modalities.
The comprehensive knowledge of a teacher model trained with trajectory, human pose, and text is distilled into a student model using only trajectory or human pose as a sole supplement.
In doing so, we separately distill the core locomotion insights from intra-agent multi-modality and inter-agent interaction.
Our generalizable framework is validated with two state-of-the-art models across three datasets on both ego-view (JRDB, SIT) and BEV-view (ETH/UCY) setups, utilizing both annotated and VLM-generated text captions.
Distilled student models show consistent improvement in all prediction metrics for both full and instantaneous observations, improving up to $\sim$13\%.
The code is available at \url{github.com/Jaewoo97/KDTF}.
\end{abstract}    
\vspace{-10pt}\section{Introduction}
\label{sec:intro}
Trajectory forecasting aims to predict the future 2D trajectory of an agent based on its historical trajectory~\cite{park2023leveraging,xu2022socialvae,chib2024pedestrian, xu2024adapting, lin2024progressive, xu2023eqmotion, gu2022stochastic, salzmann2020trajectron, gao2024multi, xu2025learning}. This predictive capability is widely applied in fields such as autonomous driving~\cite{gu2024producing,park2023leveraging,lan2023sept,zhang2024oostraj, park2024t4p, ham2023cipf, zhou2023query, liu2024laformer, cheng2023forecast, pan2024vlp, liu2024reasoning}, mobile robot navigation~\cite{salzmann2023hst, bae2024sit, saadatnejad2023social}, and surveillance systems~\cite{ahmed2018trajectory, lin2024gigatraj, pazho2024vt}. These applications depend on accurately forecasting the trajectories of human or vehicles to prevent collisions, enable smooth navigation, and improve situational awareness.

The main challenge in trajectory forecasting lies in modeling the agent's future intent in locomotion.
A clear objective on the purpose of locomotion such as a short-term (1$\sim$2 seconds) destination significantly improves the prediction accuracy \cite{jeong2024multi, lin2024progressive}.
In that sense, human agents naturally interact via visual signals to convey such intention to nearby agents during their maneuvers. 
Vehicles blink signal lights, bikers raise their arms, pedestrians rotate their torsos, all of which indicate their intent in locomotion to avoid collision with others.
Diverse applications of trajectory forecasting allow for an effortless acquisition of these visual cues.
In autonomous driving, vehicles employ various complementary sensors alongside RGB cameras as the primary tool for perception~\cite{caesar2020nuscenes, sun2020scalability}.
Mobile robots and surveillance systems also rely on RGB cameras for perception and interaction with its surrounding environment and agents~\cite{martin2021jrdb, bae2024sit}.

\begin{figure}[t!]
\includegraphics[width=\columnwidth]{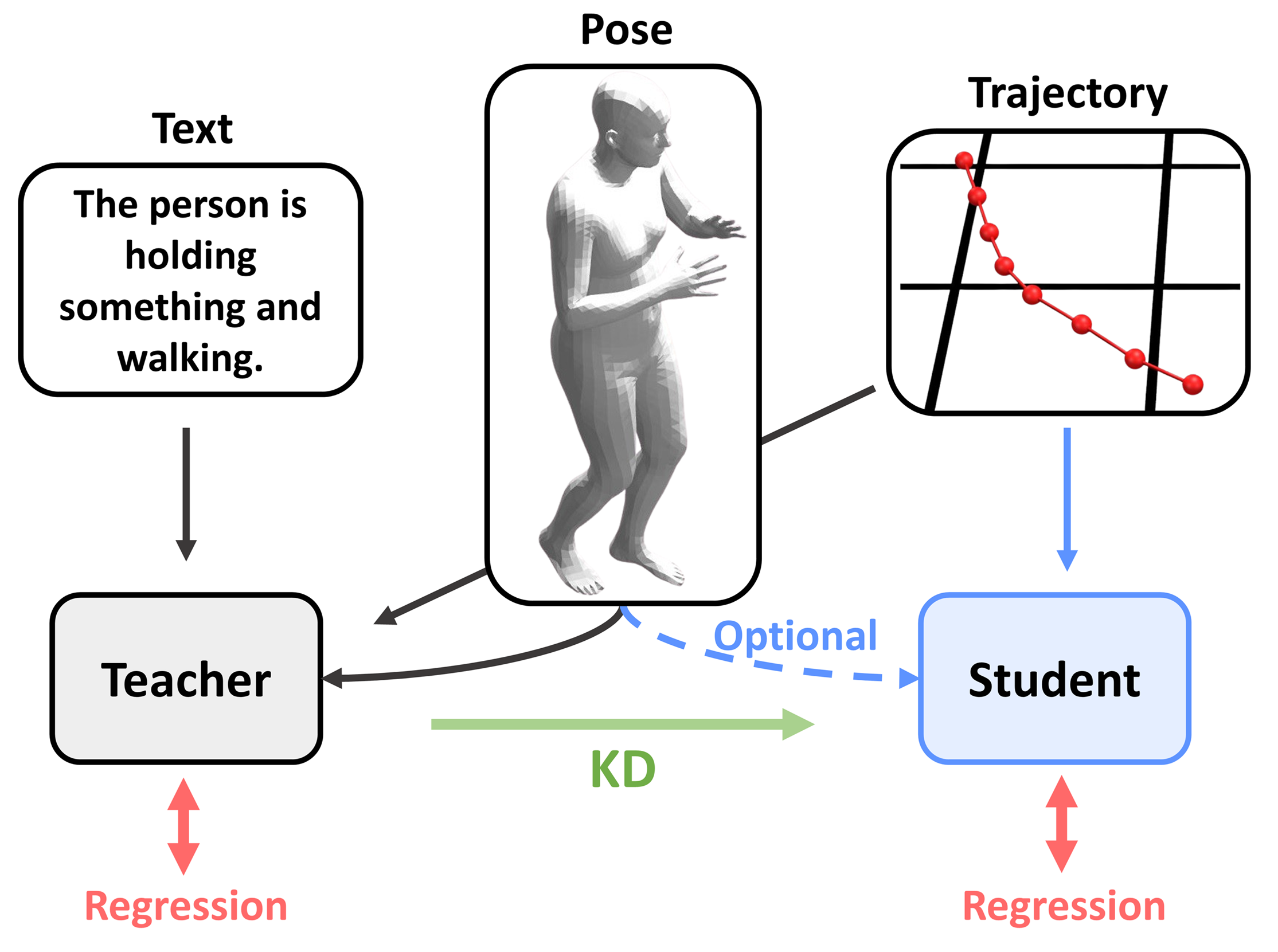}
\vspace{-17pt}
\caption{
Multi-modal data such as human pose and text greatly improve trajectory forecasting performance.
However, expensive modalities such as text are not readily available during application. 
Thus, we transfer the extensive knowledge from full modalities to a student model operating on a limited set of modalities.
}
\vspace{-10pt}
\label{fig:teaser}
\end{figure}

Recent works attempt to extract and utilize additional modalities from these visual inputs.
Extrinsics such as 2D, 3D human pose and bounding boxes have been examined to complement understanding human motion dynamics~\cite{saadatnejad2023social}.
For vehicle trajectory forecasting, vision language models (VLM) are utilized to leverage textual descriptions of the perceived environment and surrounding agents for generalization~\cite{moon2024visiontrap, bae2024can, wangtrajprompt, pan2024vlp}.
Compared to uni-modal trajectory forecasting with a sequence of mere 2D numerical coordinates, reasoning with multi-modal features allows for a more comprehensive understanding of agent's motion intent and a correspondingly accurate prediction.

Additionally, we find that textual descriptions on locomotion are essential for integrating additional modalities into a cohesive insight.
For pedestrian trajectory forecasting, text effectively bridges the domain gap between trajectory and human pose, yielding the most competent model.
This advantage manifests most distinctively in instantaneous prediction with only few input frames, as visual cues supplement the semantic context lost due to incomplete observation.

However, acquiring and processing multi-modal data substantially increases the demand in computational resources, which may exceed the capabilities of mobile systems.
In particular, acquisition of textual information requires the use of a VLM which requires significant computational resources for a reliable performance~\cite{laurenccon2024building, chen2024evlm}.

Then, how might we leverage textual insights while mitigating the computational demands associated with their acquisition?
In this regard, we propose a knowledge distillation (KD)~\cite{hinton2015distilling} framework where a student model operating with limited modalities is trained by a teacher model that utilizes the full range of modalities as shown in Fig.~\ref{fig:teaser}. 
KD resolves the computational burden, as only the affordable student model modalities are used during inference, while costly modalities are required solely during training.
Specifically, the multi-modal insight from the teacher is distilled into the modality-limited student model, enabling it to acquire language-driven comprehensive knowledge.
To fully harness the benefits of KD, we focus on pedestrian trajectory forecasting, leveraging three key aspects: i) Dynamic human pose exhibits a diverse yet coherent correlation with locomotion, ii) Pedestrian scenes span diverse indoor and outdoor environments, driving various agent-agent and agent-scene interactions, iii) Ego-view trajectory forecasting faces frequent occlusion, making it ideal for exploiting supplementary modalities.

Our framework is validated on three pedestrian trajectory forecasting datasets (JRDB~\cite{martin2021jrdb}, SIT~\cite{bae2024sit}, and ETH\cite{pellegrini2009you}/UCY~\cite{lerner2007crowds}) and with two state-of-the-art trajectory forecasting models (HiVT~\cite{zhou2022hivt} and MART~\cite{lee2024mart}).
To thoroughly assess the model's ability to integrate multi-modal data, we conduct holistic training and evaluation on both full and instantaneous observations.
The teacher model leverages trajectory, 3D human pose, and text, while the student model inputs are restricted to trajectory or 3D human pose as the sole supplement. 
While text annotations on agent's behavior and inter-agent interaction are used for JRDB, we use VLM~\cite{xu2024pllava} to extract agent descriptions on SIT.
For both datasets, a SOTA 3D pose extractor~\cite{sun2022putting} is used to extract the 3D pose.
We first train the teacher model with regression loss on both full observation and instantaneous input setups.
The student model is then distilled upon the frozen teacher model, incorporating regression loss and additional distillation loss aligning the core semantic embedding spaces.
We treat \textbf{intra}-agent and \textbf{inter}-agent embeddings as the core embeddings.
Intra-agent embeddings capture coarse locomotive implications from agent-specific modalities.
These are further encoded to account for inter-agent relationships, forming complete representations on future motion intent.
Aligning the core latent spaces across both scales enables the student model to generalize its predictive capabilities on the teacher's extensive multi-modality.
As a result, our generalized KD framework improves the distilled student model performance up to 13\% on an averaged metric. In summary, our contributions are three-fold:

- We show that textual descriptions play a crucial role in blending the modalities into a comprehensive knowledge on human motion.

- For the first time, we propose a knowledge distillation framework for human trajectory forecasting where a modality-limited student model acquires the teacher's full-modal knowledge, including VLM-generated text.

- To validate our framework, we parse two datasets that include 3D human pose and text leveraging VLM.
\section{Related works}
\subsection{Multi-modal trajectory forecasting}
Leveraging multi-modal data such as text or human pose has proven essential in recent research for both pedestrian and vehicle trajectory forecasting, particularly in environments where occlusions and complex interactions are prevalent~\cite{moon2024visiontrap, saadatnejad2023social, wangtrajprompt, pan2024vlp}.
For vehicle trajectory forecasting, recent works discovered that a language-based reasoning of motion is robust and generalizable~\cite{ma2024lampilot, cui2024receive, chen2025asynchronous}.
In doing so, leveraging VLM for perception~\cite{moon2024visiontrap, wangtrajprompt, pan2024vlp} or by directly interacting with a VLM/LLM to predict and plan future behavior has been studied~\cite{tian2024drivevlm, yuan2024rag}.
However, exploiting the robust knowledge of text on human motion has been less visited.
Instead, recent works focuses on leveraging visual cues such as bounding boxes and human poses~\cite{saadatnejad2023social, salzmann2023hst}.
To the best of our knowledge, we are first to leverage the inter-modality relationship between human pose, text, and trajectory for modeling motion intent.

\subsection{Multi-modal knowledge distillation}
While use of additional modalities greatly enhances the performance of models on a plethora of tasks~\cite{saadatnejad2023social, xu2024dmr, zhou2023unidistill, lin2024suppress}, these are often limited due to the computational limitations or requirements of bulky, expensive sensors.
Knowledge distillation serves as its solution, namely by conveying the broad knowledge of a model trained on full modalities to a student model with limited input modalities
A common protocol is to construct a competent teacher model that utilizes the full range of input modalities, upon which a student model is distilled~\cite{li2023decoupled, radevski2023multimodal, kim2025labeldistill, klingner2023x3kd, wang2020multimodal, dai2022enabling}.
Two most crucial component are defining the target knowledge to transfer and specifically tailoring a distillation method for the target modality~\cite{chae2024kd, li2024correlation, tan2023egodistill, zhang2023efficient}.
Building on these principles, we are first to propose a KD framework crafted to distill the comprehensive multi-modal knowledge on agent's motion intent for trajectory forecasting.

\begin{figure*}
    \includegraphics[width=0.99\textwidth]{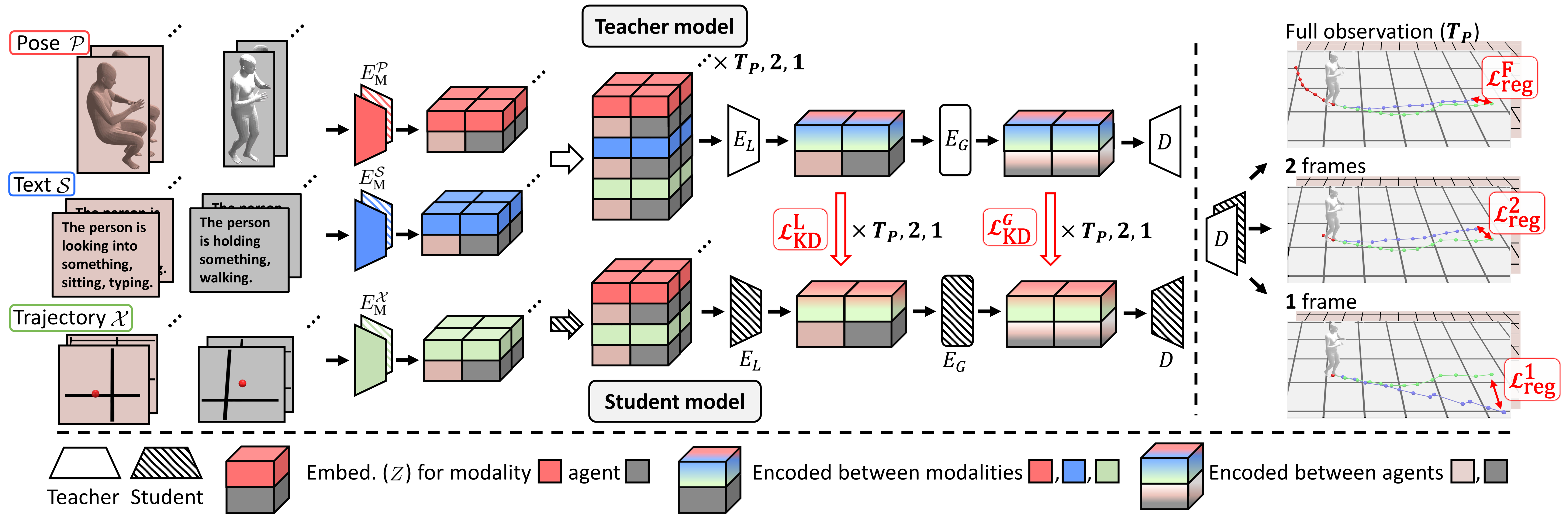} \vspace{-7pt}
    \caption{
    We first pre-train a teacher model that leverages the full range of modalities, upon which a student model with limited modalities ($\mathcal{X}+\mathcal{P}$ or $\mathcal{X}$) is distilled from scratch.
    Regression losses for three observation settings ($T_p,2,1$) are applied to both teacher and student, while additional KD losses guide the student to robustly encode intra-agent modalities ($Q$) and inter-agent interactions ($H$).
    }
    \label{fig:framework}
    \vspace{-5pt}
\end{figure*}

\subsection{Instantaneous trajectory forecasting}

trajectory forecasting in mobile systems, such as autonomous vehicles and robots, must handle unexpected agents and occlusions, relying on motion prediction from limited frames.
This poses a considerably disparate, ill-posed nature compared to conventional protocol~\cite{nathms, bae2024singulartrajectory, chib2024pedestrian}.
A key approach to addressing this challenge is incorporating past trajectories into the prediction process.
The missing past trajectory is predicted in addition to its future counterpart, thereby teaching the model to implicitly leverage the past trajectories towards predicting the future~\cite{li2023bcdiff, li2024itpnet}.
Another line of approach related to our work is utilizing knowledge distillation, where a teacher model trained on full past trajectory distills its full knowledge on its instantaneous counterpart~\cite{monti2022many, xu2024adapting}.
We take a step forward by leveraging multi-modality for knowledge distillation.
Visual and textual cues greatly supplement the model's instantaneous prediction performance, and we show that such knowledge could also be effectively conveyed to the student model.

\section{Method}
\subsection{Problem definition}
trajectory forecasting is a task of learning a mapping function between the observed 2D trajectory sequence of $N$ agents: $\mathcal{X}$ : $\left \{ \textup{\textbf{x}}_{n}^{t} \right \}_N^{-T_p:0}$, and future trajectory $\mathcal{Y}$ : $F\times\left \{ \textup{\textbf{y}}_{n}^{t} \right \}_N^{0:T_f}$, where $F$ is the number of future trajectory proposals.
We also exploit 3D human pose and text descriptions which are given in each frame, respectively notated as: $\mathcal{P}$ : $\left \{ \textup{\textbf{p}}^{t}_{n} \right \}_N^{-T_p:0}$ and 
$\mathcal{S}$ : $\left \{ \textup{\textbf{s}}^{t}_{n} \right \}_N^{-T_p:0}$.
These modalities are collectively notated as $(\mathcal{X}, \mathcal{P}, \mathcal{S})\in \mathcal{M}$.
For JRDB dataset with text annotation on agent interaction, $\mathcal{S}$ includes $s_\text{A}$ and also $s_\text{R}$, respectively for agent behavior and interaction relationship.
For the remaining method section, notations $\phi$, $\psi$ denote transformer and graph networks, respectively.

\subsection{Overall framework}
The crux of our KD framework shown in Fig.~\ref{fig:framework} lies in two aspects: i)   Constructing a competent teacher that musters a comprehensive knowledge on motion intent from three modalities: $\mathcal{X}$, $\mathcal{P}$, and $\mathcal{S}$. ii) Conveying such insight towards student models treating $\mathcal{X}$ or $\mathcal{X}+\mathcal{P}$. 
Additionally, we generalize the framework to be applicable to any regression-based model.
In this light, we re-implement two baseline models focusing on two components: a local encoder for encoding the intra-agent modalities and a global encoder for modeling the inter-agent interaction.
We distill each encoder's representation to its student counterpart, conveying insights on intra-agent intention and inter-agent interaction.

Specifically, both teacher and student models share the same design with a sole difference on the number of input modalities.
Each have their respective end-to-end models.
We first train the teacher model, followed by student model distilled on the frozen pre-trained teacher model.
While we detail each component for KD baselines HiVT~\cite{zhou2022hivt} and MART~\cite{lee2024mart}, other baselines are implemented in a similar fashion, detailed in supplementary materials.

\subsubsection{Modality embedder}
We use separate encoders $E_\text{M}^\mathcal{X}, E_\text{M}^\mathcal{P}, E_\text{M}^\mathcal{S}$ to encode each modality.
Following conventional approaches, trajectory embedding is obtained using a MLP.
Depending on the model, normalized global position and temporal difference of positions are both encoded or selectively encoded as trajectory embedding: ${z_{x}}^t_n = E^{\mathcal{X}}_{M}(\mathcal{X})$.
In addition, $E^{\mathcal{X}}_{M}$ also encodes heading angle for trajectory-only student models to acquire context on instantaneous forecasting given 1 frame.
For text, we use a pre-trained TinyBERT encoder~\cite{jiao2019tinybert} to obtain the sentences' class tokens as text embeddings, thereby introducing semantic context: ${z_{s}}^t_n = E^{\mathcal{S}}_{M}(\mathcal{S})$.
For human pose, we use an MLP to encode the SMPL theta parameters into pose embedding: ${z_{p}}^t_n = E^{\mathcal{P}}_{M}(\mathcal{P})$. 
We use simple MLP networks for embedders to focus on the effects of cross-modality fusion via these embeddings, not the processing of obtaining the embeddings itself.
As a result, we obtain modality-specific embeddings for $\mathcal{X}$, $\mathcal{P}$, and $\mathcal{S}$ as:

\begin{equation}
\begin{aligned}
\textup{$Z=\{{z_{x}}^t_n, {z_{p}}^t_n, {z_{s_A}}^t_n, {z_{s_R}}^t_n\}\in\mathbb{R}^{N_A\times T_p\times D}$} \\
\end{aligned}
\end{equation}


\subsubsection{Local encoder}
The local encoder $E_\text{L}$ fuses modalities in both the modality ($\mathcal{M}$) and temporal ($T_p$) dimensions, capturing a detailed representation of each agent's behavior. 
Since all baseline models are transformer-based, we use cross-attention to fuse modalities. 
MART originally processed $\mathcal{X}$ with an MLP, which we replace with a transformer module $E_\text{L}$ to encode both $\mathcal{M}$ and $T_p$ axes via holistic attention. 
A class token $\overline{q_n}$ is added with the modalities to acquire a representative embedding for each agent, similar to BERT~\cite{devlin2018bert}.

\begin{equation}
\begin{aligned}
\textup{$q_n = E_\text{L}^{\text{MART}}(Z) = \phi_{\mathcal{M}, T_p}(\overline{q_n},z_x, z_p, z_s)$} \\
\end{aligned}
\end{equation}

For HiVT, modality and temporal dimensions are processed sequentially to support graph-based interaction modeling per frame. 
Unlike standard transformer encoders that apply attention across embeddings simultaneously, HiVT’s graph-based approach models relationships between each agent and its neighbors individually. 
Prior to encoding, HiVT rotates each neighbor’s trajectory to the ego vehicle’s heading vector coordinates, achieving rotation-invariant embeddings.
Such design aims to encode a more thorough agent representation by adaptively handling each inter-agent relationship.
We extend this design to incorporate human pose for neighbors within a threshold 
$\tau$, addressing pose correlations in a rotation-invariant manner to more effectively capture motion intents from subtle interactions and nuances reflected in pose.
    
For each frame, agent modalities $(\mathcal{X}^t$, $\mathcal{P}^t$, $\mathcal{S}^t)_i$ and the neighbor modalities $(\mathcal{X}^t$, $\mathcal{P}^t)_j$ are encoded as follows:

\begin{equation}
\begin{aligned}
\{z_{x},z_{p},z_{s}\}_i&=\mathcal{E}(\{x^t, p^t, s^t\}_i) \\ 
\{z_{x},z_{p}\}_i&=\mathcal{E}(\{R_{ji}x^t, R_{ji}p^t\}_j)
\end{aligned}
\end{equation}

\begin{equation}
\begin{aligned}
q_n^t &= E_\text{L}^{\text{HiVT}}(\mathcal{M}) \\
&= \psi_{\mathcal{M}} \left( \left[ (z_x, z_p, z_s)_i, (z_x, z_p, z_s)_j, (v_{ji})_e \right] \right)
\end{aligned}
\end{equation}

$\psi_{\mathcal{M}}[()_i,()_j,()_e]$ denotes a graph operation encoding the ego ($i$) feature via cross attention with neighbor ($j$) and edge ($e$) features.
$v_{ji}$ denotes a vector from $j$ to $i$ position at time $t=0$.
$R_{ji}$ represents a rotation matrix that maps the neighbor trajectory from global coordinates to agent's heading vector coordinates.
Then, the temporal domain is encoded by a transformer as in $E^{\text{MART}}_\text{L}$.
The resulting embedding $q_n\in\mathbb{R}^N\times D$ comprehensively represents the coarse motion intents inferred from multi-modal input.

\subsubsection{Global encoder}
The global encoder $E_\text{G}$ subsequently attends to inter-agent interaction among $q_n$, forming complete implications on socially compliant locomotion.
MART employs a group prediction-based transformer model to incorporate inter-agent interactions, a design that aligns well with the integration of additional modalities. 
Leveraging these modalities enhances the robustness of inter-agent relationship inferences, thereby supporting more reliable group assignments.

\begin{equation}
\begin{aligned}
\textup{$H = E_\text{G}^{\text{MART}}(Q) = \phi_{N}(Q)$}\\
\end{aligned}
\end{equation}

For HiVT, we again exploit the advantages of a graph-based model by respectively leveraging text for each interaction.
JRDB explicitly includes text annotations on the relationship between agents, such as ``They are standing together and having conversation."
These text embeddings are used to encode edge attributes along with a spatial cue of 2D vector at t=0 across agent-neighbor pairs as follows:

\begin{equation}
\begin{aligned}
H &= E_\text{G}^{\text{HiVT}}(Q) &= \psi_{N} \left( \left[ (q_n)_i, (q_n)_j, (v_{ji}, s_{R,ji}) \right] \right)
\end{aligned}
\end{equation}

\subsubsection{Decoder}
We use a standard MLP decoder $D$ that maps the global encoder output $H\in\mathbb{R}^{N_A\times D}$ to future trajectory forecasts with multiple proposals, $\mathcal{Y}\in\mathbb{R}^{F\times T_f\times 2}$.  

\subsection{Training objectives}
\subsubsection{Regression loss}
We regress each model with their corresponding default losses for both teacher and student models: L2 for MART and Negative Log Likelihood for HiVT.
Compared to a conventional protocol of only training with full past observation, we also train the models with instantaneous observations (1 or 2 frames) to thoroughly assess the model’s ability in integrating multi-modality for challenging situations.
All observations other than the last 2 or 1 frames are padded for the instantaneous prediction setting.
These losses are notated as $L^\text{F}_{\text{reg}}$, $L^\text{2}_{\text{reg}}$, $L^\text{1}_{\text{reg}}$, each corresponding to loss with full, 2 frames, and 1 frame of observation.
The teacher model is first trained via these regression losses.

\begin{figure*}
    \includegraphics[width=0.99\textwidth]{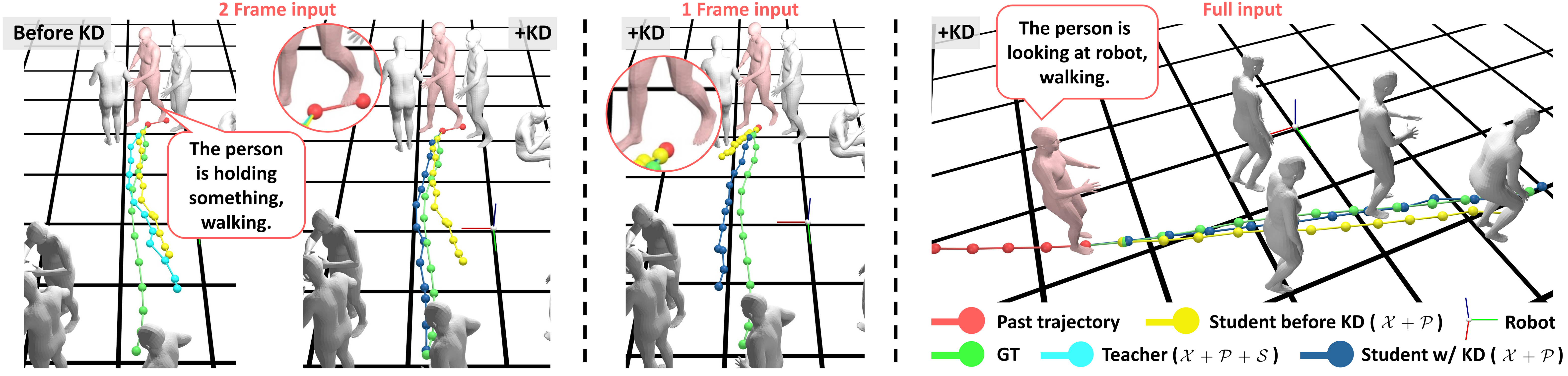} \vspace{-7pt}
    \caption{
    Qualitative results on JRDB with $\mathcal{X}+\mathcal{P}$ HiVT model.
    The model outperforms its baseline counterpart with KD.
    Improved accuracy is demonstrated on all instantaneous and full observations.
    Origin denotes robot position, and bubbles represent text annotations.
    }
    \label{fig:qual}
    \vspace{-5pt}
\end{figure*}

\subsubsection{Knowledge distillation loss}
We distill the two core knowledges in trajectory forecasting: 1. $Q$, the fusion of intra-agent multi-modalities into a coarse motion intent, 2. $H$, the refined implications on future locomotion incorporating agent interactions.
In doing so, we apply KL divergence between the teacher ($Q_\mathcal{T}$, $H_\mathcal{T}$) and student latents ($Q_\mathcal{S}$, $H_\mathcal{S}$) to align their distributions.
\begin{equation}
\begin{aligned}
\mathcal{L}_{\text{KD}} &= \mathcal{L}_{\text{KD}}^{\text{L}} + \mathcal{L}_{\text{KD}}^{\text{G}} \\&= \mathcal{L}_{\text{KL}}(Q_\mathcal{T} \| Q_\mathcal{S}) + \mathcal{L}_{\text{KL}}(H_\mathcal{T} \| H_\mathcal{S})
\end{aligned}
\end{equation}
For HiVT, we add an additional regularization term for training stability and use cosine similarity for KD:
\begin{equation}
\begin{aligned}
\mathcal{L}_{\text{KD}}^\text{L} &= \lambda_{\text{cos}} \mathcal{L}_{\text{cos}}(Q_\mathcal{T},Q_\mathcal{S}) + \mathcal{L}_{\text{KL}}(\mathcal{N} \| Q_\mathcal{S}) \\\mathcal{L}_{\text{KD}}^\text{G} &= \lambda_{\text{cos}} \mathcal{L}_{\text{cos}}(H_\mathcal{T},H_\mathcal{S}) + \mathcal{L}_{\text{KL}}(\mathcal{N} \| H_\mathcal{S})
\end{aligned}
\end{equation}

Again, $\mathcal{L}_{\text{KL}}$ is computed for all three types of observations: $\mathcal{L}_{\text{KD}}^{\text{F}}$, $\mathcal{L}_{\text{KD}}^{\text{2}}$, $\mathcal{L}_{\text{KD}}^{\text{1}}$, each denoting full, 2, and 1 frame of observation.
The overall loss for a student model is:

\begin{equation}
\begin{aligned}
\mathcal{L} = \lambda_{\text{reg}} L^\text{F}_{\text{reg}} + L^\text{2}_{\text{reg}} + L^\text{1}_{\text{reg}} + \mathcal{L}_{\text{KD}}^{\text{F}} +\mathcal{L}_{\text{KD}}^{\text{2}} + \mathcal{L}_{\text{KD}}^{\text{1}}
\end{aligned}
\end{equation}

Where $\lambda$ denotes scaling factor: 0.5 for $\lambda_{\text{cos}}$ and 3 for $\lambda_{\text{reg}}$.
These losses guide the student to predict future trajectories while aligning its core latents with the teacher. 
By mimicking the teacher's behavior while learning to forecast the future, the student is encouraged to operate as robustly or even surpass the teacher via improved generalization.

\begin{table*}[]
\caption{Prediction results on JRDB with multimodal inputs: human pose $\mathcal{P}$ and text $\mathcal{S}$. \textbf{Bold} denotes best. Lower is better.}
\vspace{-7pt}
\resizebox{\textwidth}{!}{%
\begin{tabular}{|c|ccccccc|ccccccc|}
\hline
\multirow{2}{*}{$\mathcal{M}$} & \multicolumn{7}{c|}{HiVT} & \multicolumn{7}{c|}{MART} \\ \cline{2-15} 
 & ADE & \text{ADE}\textsubscript{2} & \text{ADE}\textsubscript{1} & FDE & \text{FDE}\textsubscript{2} & \multicolumn{1}{c|}{\text{FDE}\textsubscript{1}} & Ave. +\%\tablefootnote{Average of \% improvement for each metric. Same for all Ave. +\% notations in the remainder of the paper.} & ADE & \text{ADE}\textsubscript{2} & \text{ADE}\textsubscript{1} & FDE & \text{FDE}\textsubscript{2} & \multicolumn{1}{c|}{\text{FDE}\textsubscript{1}} & Ave. +\% \\ \hline
$\mathcal{X}$ & 0.221 & 0.240 & 0.342 & 0.432 & 0.467 & \multicolumn{1}{c|}{0.632} & - & 0.286 & 0.275 & 0.395 & 0.545 & 0.526 & \multicolumn{1}{c|}{0.753} & - \\
$\mathcal{X} + \mathcal{P}$ & 0.229 & 0.242 & 0.364 & 0.441 & 0.465 & \multicolumn{1}{c|}{0.659} & -2.84 & 0.287 & 0.282 & 0.366 & 0.543 & 0.538 & \multicolumn{1}{c|}{0.682} & +2.02 \\
$\mathcal{X} + \mathcal{S}$ & 0.224 & 0.230 & 0.285 & 0.432 & 0.444 & \multicolumn{1}{c|}{0.540} & +6.53 & 0.261 & 0.265 & 0.301 & 0.496 & 0.506 & \multicolumn{1}{c|}{0.564} & +12.41 \\
$\mathcal{X} + \mathcal{P} + \mathcal{S}$ & \textbf{0.220} & \textbf{0.227} & \textbf{0.280} & \textbf{0.423} & \textbf{0.435} & \multicolumn{1}{c|}{\textbf{0.521}} & \textbf{+8.38} & \textbf{0.258} & \textbf{0.256} & \textbf{0.289} & \textbf{0.491} & \textbf{0.486} & \multicolumn{1}{c|}{\textbf{0.538}} & \textbf{+14.98} \\ \hline
\end{tabular}
}
\vspace{-8pt}
\label{table:main_multimodal}
\end{table*}

\section{Experiment}
\subsection{Dataset}
We test our model on three datasets: JRDB~\cite{martin2021jrdb}, SIT~\cite{bae2024sit}, and ETH~\cite{pellegrini2009you}/UCY~\cite{lerner2007crowds}.
JRDB and SIT are the largest datasets for human tracking and prediction from a robot perspecitve, suitable for developing our framework with their affluent visual features.
In addition, traditional ETH/UCY benchmark is also tested upon to show the versatility of our approach even on limited visual features.

For JRDB and SIT, we utilize 3D human pose and text captions as an additional modality of visual feature.
We extract the SMPL theta parameters of 3D human pose using~\cite{sun2022putting} for each visible agent.
Unlike previous works~\cite{salzmann2023hst, saadatnejad2023social, jeong2024multi}, we use SMPL representation for human pose for improved generalizability as discussed in supplementary materials.
For text captions, annotated text descriptions~\cite{jahangard2024jrdb} are parsed to describe intra-agent behavior and inter-agent interaction for JRDB dataset.
For SIT dataset, we use PLLaVa~\cite{xu2024pllava} to caption agent descriptions from consecutive frames of cropped images of agents, using prompt ``What is the person doing?".
As JRDB dataset includes more diverse scenes and agent behaviors, with up to 38 agents and a 2$\times$ longer duration, we treat JRDB as our primary dataset for KD and the ablation studies.
Both JRDB and SIT are parsed with 2.5FPS to predict 12 frames given 8 frames, matching the ETH/UCY dataset configuration.

For ETH/UCY dataset, its BEV-view and low resolution present disparate limitations compared to JRDB and SIT datasets. 
While human pose has been used as visual feature for JRDB and SIT, neither 3D or 2D pose are able to be accurately extracted due to low visibility of human pose.
Therefore, we resort to 2D image features of cropped images from pretrained CLIP image encoder~\cite{radford2021learning} as visual features. 
For text captions, VLM models were incompetent in accurately captioning agent behavior for BEV-views.
Instead, we implement a rule-based approach to generate text captions that describe the map context, such as ``There is an obstacle in the right."
Further details regarding datasets are scrutinized in the supplementary materials.

\subsection{Metrics}
We use the widely used metrics for trajectory forecasting, Average Displacement Error (ADE) and Final Displacement Error (FDE).
For all experimentation, we predict 6 modes ($F$) of future trajectories to assess the model's competence in predicting a few accurate trajectories rather than over-populating similar trajectories.
We also evaluate performance on instantaneous forecasting, where only the most recent one or two frames of the past trajectory are given, denoted as $ADE_1$ and $ADE_2$.

\subsection{Baselines}
We choose the following works as baselines and additionally perform KD on underlined models. We precisely modify each model to have similar number of parameters. Further details are found in the supplementary materials.\newline
\textbf{\underline{HiVT}}~\cite{zhou2022hivt} is a graph-based network designed with attention-based computation. 
We choose HiVT for KD due to its SOTA performance and flexibility in incorporating additional modalities across both ego and interaction dimensions, enabled by its graph-based architecture.\newline
\textbf{\underline{MART}}~\cite{lee2024mart} is a transformer-based network and is one of most recent SOTA model for pedestrian trajectory forecasting. 
We choose MART for KD to demonstrate generalizing our KD framework on any transformer-based models, even ones not specifically designed to handle multi-modality.\newline
\textbf{ST}~\cite{saadatnejad2023social} (socialTransmotion) is a transformer-based model for human trajectory forecasting designed to handle additional human modalities such as 2D, 3D pose and bounding boxes. This aligns with the motivations of our framework and makes it a suitable candidate for comparison.\newline
\textbf{LED}~\cite{mao2023leapfrog} is a transformer-based diffusion model achieving SOTA performance on pedestrian trajectory forecasting, chosen to compare our work on various types of learning-based models including diffusion models.

\begin{table}[]
\caption{KD on ETH/UCY (average) with HiVT student model using $\mathcal{X}$. \% measures improvement with KD. \textbf{Bold} denotes best. Grey background highlights results with our KD framework.}
\vspace{-7pt}
\resizebox{\columnwidth}{!}{%
\begin{tabular}{|c|c|cccccc|c|}
\hline
Model & \begin{tabular}[c]{@{}c@{}}w/ \\ KD\end{tabular} & ADE & \multicolumn{1}{l}{\text{ADE}\textsubscript{2}} & \text{ADE}\textsubscript{1} & FDE & \multicolumn{1}{l}{\text{FDE}\textsubscript{2}} & \text{FDE}\textsubscript{1}& \begin{tabular}[c]{@{}c@{}}Ave.\\ + \%\end{tabular} \\ \hline
LED & - & 0.312 & 0.334 & 0.600 & 1.008 & 1.017 & 1.427 & - \\ \hline
ST & - & 0.319 & 0.319 & 0.605 & 0.853 & 0.855 & 1.224 & - \\ \hline
 &  & 0.364 & 0.369 & 0.571 & 0.852 & 0.862 & 1.166 &  \\
\multirow{-2}{*}{MART} & \cellcolor[HTML]{DADADA}\checkmark & \cellcolor[HTML]{DADADA}0.347 & \cellcolor[HTML]{DADADA}0.353 & \cellcolor[HTML]{DADADA}0.553 & \cellcolor[HTML]{DADADA}0.807 & \cellcolor[HTML]{DADADA}0.817 & \cellcolor[HTML]{DADADA}1.141 & \multirow{-2}{*}{\textbf{+3.80}} \\ \hline
 &  & 0.314 & 0.320 & 0.555 & 0.718 & \textbf{0.726} & 1.132 &  \\
\multirow{-2}{*}{HiVT} & \cellcolor[HTML]{DADADA}\checkmark & \cellcolor[HTML]{DADADA}\textbf{0.304} & \cellcolor[HTML]{DADADA}\textbf{0.317} & \cellcolor[HTML]{DADADA}\textbf{0.546} & \cellcolor[HTML]{DADADA}\textbf{0.698} & \cellcolor[HTML]{DADADA}0.727 & \cellcolor[HTML]{DADADA}\textbf{1.122} & \multirow{-2}{*}{+1.55} \\ \hline
\end{tabular}
}
\label{table:ethucy}
\vspace{-14pt}
\end{table}
\begin{table*}[]
\caption{KD results on JRDB and SIT dataset. Full modality $\mathcal{X}+\mathcal{P}+\mathcal{S}$ model is chosen as teacher model. KD is performed on student models using modality $\mathcal{X}$ and $\mathcal{X}+\mathcal{P}$. \% measures improvement with KD. \textbf{Bold} represents best performance.}
\vspace{-7pt}
\resizebox{\textwidth}{!}{%
\begin{tabular}{|c|c|c|ccccccc|ccccccc|}
\hline
 &  &  & \multicolumn{7}{c|}{JRDB dataset} & \multicolumn{7}{c|}{SIT dataset} \\ \cline{4-17} 
\multirow{-2}{*}{Modality} & \multirow{-2}{*}{Model} & \multirow{-2}{*}{\begin{tabular}[c]{@{}c@{}}w/\\ KD\end{tabular}} & ADE & \text{ADE}\textsubscript{2} & \text{ADE}\textsubscript{1} & FDE & \text{FDE}\textsubscript{2} & \multicolumn{1}{c|}{\text{FDE}\textsubscript{1}} & Avg. +\% & ADE & \text{ADE}\textsubscript{2} & \text{ADE}\textsubscript{1} & FDE & \text{FDE}\textsubscript{2} & \multicolumn{1}{c|}{\text{FDE}\textsubscript{1}} & Avg. +\% \\ \hline
 & LED &  & 0.358 & 0.426 & 0.447 & 0.760 & 0.897 & \multicolumn{1}{c|}{0.934} & - & 0.479 & 0.507 & 0.546 & 1.045 & 1.057 & \multicolumn{1}{c|}{1.093} & - \\ \cline{2-17} 
 & ST &  & 0.324 & 0.337 & 0.427 & 0.614 & 0.636 & \multicolumn{1}{c|}{0.780} & - & 0.531 & 0.544 & 0.699 & 1.037 & 1.065 & \multicolumn{1}{c|}{1.310} & - \\ \cline{2-17} 
 &  &  & 0.286 & 0.275 & 0.395 & 0.545 & 0.526 & \multicolumn{1}{c|}{0.753} &  & 0.468 & 0.483 & 0.556 & 0.909 & 0.937 & \multicolumn{1}{c|}{1.073} &  \\
 & \multirow{-2}{*}{MART} & \cellcolor[HTML]{DADADA}\checkmark & \cellcolor[HTML]{DADADA}0.259 & \cellcolor[HTML]{DADADA}0.262 & \cellcolor[HTML]{DADADA}0.366 & \cellcolor[HTML]{DADADA}0.495 & \cellcolor[HTML]{DADADA}0.496 & \multicolumn{1}{c|}{\cellcolor[HTML]{DADADA}0.684} & \multirow{-2}{*}{\textbf{+7.61}} & \cellcolor[HTML]{DADADA}0.441 & \cellcolor[HTML]{DADADA}0.465 & \cellcolor[HTML]{DADADA}0.570 & \cellcolor[HTML]{DADADA}0.832 & \cellcolor[HTML]{DADADA}0.885 & \multicolumn{1}{c|}{\cellcolor[HTML]{DADADA}1.090} & \multirow{-2}{*}{+3.25} \\ \cline{2-17} 
 &  &  & 0.221 & 0.240 & 0.342 & \textbf{0.432} & 0.467 & \multicolumn{1}{c|}{0.632} &  & 0.455 & 0.457 & 0.526 & 0.869 & 0.884 & \multicolumn{1}{c|}{0.988} &  \\
\multirow{-6}{*}{$\mathcal{X}$} & \multirow{-2}{*}{HiVT} & \cellcolor[HTML]{DADADA}\checkmark & \cellcolor[HTML]{DADADA}\textbf{0.220} & \cellcolor[HTML]{DADADA}\textbf{0.232} & \cellcolor[HTML]{DADADA}\textbf{0.326} & \cellcolor[HTML]{DADADA}0.438 & \cellcolor[HTML]{DADADA}\textbf{0.453} & \multicolumn{1}{c|}{\cellcolor[HTML]{DADADA}\textbf{0.604}} & \multirow{-2}{*}{+2.38} & \cellcolor[HTML]{DADADA}\textbf{0.404} & \cellcolor[HTML]{DADADA}\textbf{0.427} & \cellcolor[HTML]{DADADA}\textbf{0.503} & \cellcolor[HTML]{DADADA}\textbf{0.799} & \cellcolor[HTML]{DADADA}\textbf{0.833} & \multicolumn{1}{c|}{\cellcolor[HTML]{DADADA}\textbf{0.965}} & \multirow{-2}{*}{\textbf{+6.37}} \\ \hline
 & LED &  & 0.340 & 0.358 & 0.446 & 0.715 & 0.822 & \multicolumn{1}{c|}{0.949} & - & 0.469 & 0.491 & 0.548 & 1.051 & 1.045 & \multicolumn{1}{c|}{1.085} & - \\ \cline{2-17} 
 & ST &  & 0.345 & 0.362 & 0.437 & 0.637 & 0.663 & \multicolumn{1}{c|}{0.782} & - & 0.541 & 0.576 & 0.751 & 1.046 & 1.119 & \multicolumn{1}{c|}{1.410} & - \\ \cline{2-17} 
 &  &  & 0.287 & 0.282 & 0.366 & 0.543 & 0.538 & \multicolumn{1}{c|}{0.682} &  & 0.473 & 0.484 & 0.574 & 0.909 & 0.937 & \multicolumn{1}{c|}{1.119} &  \\
 & \multirow{-2}{*}{MART} & \cellcolor[HTML]{DADADA}\checkmark & \cellcolor[HTML]{DADADA}0.266 & \cellcolor[HTML]{DADADA}0.261 & \cellcolor[HTML]{DADADA}0.337 & \cellcolor[HTML]{DADADA}0.519 & \cellcolor[HTML]{DADADA}0.507 & \multicolumn{1}{c|}{\cellcolor[HTML]{DADADA}0.633} & \multirow{-2}{*}{\textbf{+6.63}} & \cellcolor[HTML]{DADADA}0.448 & \cellcolor[HTML]{DADADA}0.465 & \cellcolor[HTML]{DADADA}0.550 & \cellcolor[HTML]{DADADA}0.849 & \cellcolor[HTML]{DADADA}0.883 & \multicolumn{1}{c|}{\cellcolor[HTML]{DADADA}1.059} & \multirow{-2}{*}{+5.19} \\ \cline{2-17} 
 &  &  & \textbf{0.229} & 0.242 & 0.364 & \textbf{0.441} & 0.465 & \multicolumn{1}{c|}{0.659} &  & 0.518 & 0.519 & 0.531 & 0.979 & 0.979 & \multicolumn{1}{c|}{1.006} &  \\
\multirow{-6}{*}{$\mathcal{X}$ + $\mathcal{P}$} & \multirow{-2}{*}{HIVT} & \cellcolor[HTML]{DADADA}\checkmark & \cellcolor[HTML]{DADADA}0.232 & \cellcolor[HTML]{DADADA}\textbf{0.239} & \cellcolor[HTML]{DADADA}\textbf{0.308} & \cellcolor[HTML]{DADADA}0.445 & \cellcolor[HTML]{DADADA}\textbf{0.464} & \multicolumn{1}{c|}{\cellcolor[HTML]{DADADA}\textbf{0.560}} & \multirow{-2}{*}{+4.98} & \cellcolor[HTML]{DADADA}\textbf{0.414} & \cellcolor[HTML]{DADADA}\textbf{0.444} & \cellcolor[HTML]{DADADA}\textbf{0.500} & \cellcolor[HTML]{DADADA}\textbf{0.789} & \cellcolor[HTML]{DADADA}\textbf{0.853} & \multicolumn{1}{c|}{\cellcolor[HTML]{DADADA}\textbf{0.951}} & \multirow{-2}{*}{\textbf{+13.03}} \\ \hline
\end{tabular}
}
\label{table:main_jrdbsit_alltimesteploss}
\vspace{-10pt}
\end{table*}
\section{Results}
\subsection{Quantitative results}
\subsubsection{Multi-modal teacher model}
\label{sec:multimodal}
As with any KD framework, we start off by constructing a competent teacher model via leveraging diverse modalities.
Table \ref{table:main_multimodal} demonstrates the performance improvement with the use of additional modalities. 
Both HiVT and MART gradually improves over all metrics with more modality supplements, achieving up to 14.98\% improvement in performance.
Such significant improvement demonstrates that $\mathcal{P}$ and $\mathcal{S}$ greatly aids understanding motion intent and interpreting the corresponding future motion.

Specifically, $\mathcal{S}$ plays a key role in building robust knowledge.
For both models, the addition of $\mathcal{P}$ results in limited improvement, primarily due to the noise introduced during pose extraction.
In contrast, all metrics considerably improve with $\mathcal{S}$, reflecting the strong correlation between human motion and language. 
Leveraging this correlation fully exploits the potential of human pose as shown by further improvement from $\mathcal{X}+\mathcal{P}$ to $\mathcal{X}+\mathcal{P}+\mathcal{S}$, even when $\mathcal{X}$ to $\mathcal{X}+\mathcal{P}$ has shown degradation for HiVT.
This shows that language bridges the domain gap between trajectory and noisy human pose, forming a more comprehensive understanding of motion intent.
Use of modalities is particularly beneficial on instantaneous settings, as additional modalities supplement the missing context of instantaneous trajectory, improving up to 27\% for $\text{ADE}_1$ on MART from $\mathcal{X}$ to $\mathcal{X}+\mathcal{P}+\mathcal{S}$.

\subsubsection{Multi-modal knowledge distillation}
\label{sec:multimodal KD}
Full modality $\mathcal{X}+\mathcal{P}+\mathcal{S}$ model is used as a teacher for all datasets.
Table~\ref{table:ethucy} and \ref{table:main_jrdbsit_alltimesteploss} show the improvements made with KD.
For ego-view datasets JRDB and SIT, we choose Student model using $\mathcal{X}$ and $\mathcal{X}+\mathcal{P}$. 
A consistent improvement over most metrics on all settings is observed, improving up to 13\% on SIT dataset.
Notably, the performance of both modality type students improves with KD, even when only $\mathcal{X}$ is used as input modality.
This improvement shows the unlocking of the potential of numerical trajectories to grasp the semantic context-based motion intent from the knowledge of other modalities.

For JRDB dataset which contains considerable amount of training data, HiVT's vanilla $\mathcal{X}$ model performs considerably well with full $T_F$ input.
In such case, the advantages of KD is spotlighted on instantaneous prediction where the improvement averages around 5$\sim$10\%.
$\mathcal{X+P}$ for HiVT also shows a similar tendency, showing significant improvement on instantaneous prediction.
Repeated improvement on $\mathcal{X+P}$ again showcases the maximal exploitation of human pose, guided by multi-modal knowledge.

Compared to the JRDB dataset, SIT dataset is a smaller dataset with less number of agents and shorter total duration.
For such scarce setting, the base model suffered to establish a strong understanding between $\mathcal{X}$ and $\mathcal{P}$ and resulted in inferior performance even with $\mathcal{P}$ as additional modality.
The strength of KD is particularly evident in this setting, showing the greatest improvement of 13\%.
consistent improvement with KD on $\mathcal{X}+\mathcal{P}$ confirms the crucial role of text in establishing a firm understanding on the correlation between human pose and locomotion intent via KD.

Our KD framework's competence is also manifested on ETH/UCY dataset as shown in Tab.~\ref{table:ethucy}. 
As mentioned in experiment section, cropped 2D image feature is used in place of human pose and text describes the nearby obstacles.
Since validation is performed on a different scene, use of image features does not inherently guarantee any generalizability.
On the other hand, text contains generalizable map context information in unified format such as ``There is an obstacle on the right of the person."
This provides hints on reducing the agent motion intent's degree of freedom.
The teacher model learns to leverage these hints for forecasting, especially improving their instantaneous prediction performance.
Upon KD, $\mathcal{X}$ student models were able to learn such knowledge and improved between 1$\sim$ 4\%.
Consistent improvement across all ego, BEV-view datasets demonstrate the generalizability of our multi-modal KD framework.

\begin{table}[]
\caption{Ablation on KD components, namely $E_L$ and $E_G$ outputs. Models use $\mathcal{X}+\mathcal{P}$ modalities. Experimented on JRDB.}
\vspace{-7pt}
\resizebox{\columnwidth}{!}{%
\begin{tabular}{|ccccccccc|}
\hline
$\mathcal{L}_{\text{KD}}^\text{L}$ & \multicolumn{1}{c|}{$\mathcal{L}_{\text{KD}}^\text{G}$} & ADE & \text{ADE}\textsubscript{2} & \text{ADE}\textsubscript{1} & FDE & \text{FDE}\textsubscript{2} & \multicolumn{1}{c|}{\text{FDE}\textsubscript{1}} & Avg. +\% \\ \hline
\multicolumn{9}{|c|}{HiVT} \\ \hline
\multicolumn{1}{|l}{} & \multicolumn{1}{l|}{} & 0.229 & 0.242 & 0.364 & 0.441 & 0.465 & \multicolumn{1}{c|}{0.659} & - \\
\multicolumn{1}{|l}{} & \multicolumn{1}{c|}{\checkmark} & 0.228 & 0.244 & 0.352 & 0.447 & 0.472 & \multicolumn{1}{c|}{0.647} & +0.26 \\
\checkmark & \multicolumn{1}{l|}{} & 0.228 & 0.241 & 0.327 & 0.445 & 0.468 & \multicolumn{1}{c|}{0.601} & +3.09 \\
\rowcolor[HTML]{DADADA} 
\checkmark & \multicolumn{1}{c|}{\cellcolor[HTML]{DADADA}\checkmark} & \textbf{0.232} & \textbf{0.239} & \textbf{0.308} & \textbf{0.445} & \textbf{0.464} & \multicolumn{1}{c|}{\cellcolor[HTML]{DADADA}\textbf{0.560}} & \textbf{+4.98} \\ \hline
\multicolumn{9}{|c|}{MART} \\ \hline
\multicolumn{1}{|l}{} & \multicolumn{1}{l|}{} & 0.287 & 0.282 & 0.366 & 0.543 & 0.538 & \multicolumn{1}{c|}{0.682} & - \\
\multicolumn{1}{|l}{} & \multicolumn{1}{c|}{\checkmark} & 0.274 & 0.272 & 0.380 & 0.520 & 0.514 & \multicolumn{1}{c|}{0.711} & +1.49 \\
\checkmark & \multicolumn{1}{l|}{} & 0.275 & 0.271 & 0.366 & 0.529 & 0.520 & \multicolumn{1}{c|}{0.683} & +2.33 \\
\rowcolor[HTML]{DADADA} 
\checkmark & \multicolumn{1}{c|}{\cellcolor[HTML]{DADADA}\checkmark} & \textbf{0.266} & \textbf{0.261} & \textbf{0.337} & \textbf{0.519} & \textbf{0.507} & \multicolumn{1}{c|}{\cellcolor[HTML]{DADADA}\textbf{0.633}} & \textbf{+6.63} \\ \hline
\end{tabular}
}
\vspace{-10pt}
\label{table:ablation_KD_components}
\end{table}

\subsection{Qualitative results}
Figure~\ref{fig:qual} shows predictions made with both instantaneous and full observations on JRDB dataset.
The leftmost two figures visualize predictions before and after KD.
Surprisingly, the student model outperforms the teacher model after KD on prediction with 2 frame input. 
This shows that the student not only mimics the teacher model, but generalizes itself to the task based on the comprehensive knowledge of the teacher.
In the extreme of a single frame input, the student before KD collides into a nearby agent.
After KD, however, the model accurately predicts a collision-free trajectory.
Notably, the model successfully infers the direction of avoidance, demonstrating a solid understanding of pose orientation after KD.
Overall, student model shows distinct improvements across all input settings with KD. More qualitative results are found in the supplementary materials.

\subsection{Ablation studies}
\label{sec:ablation}
Herein, we delve into deeper questions regarding the dual foundation of our KD framework: \ref{sec:multimodal}, \ref{sec:multimodal KD}. Further ablations are found in the supplementary materials.\\
\textbf{How do we ensure an effective knowledge distillation?}\\
The common core of various KD frameworks lies in aligning the latent space across different modalities.
Latent spaces sharing a comparable semantic meaning is precisely aligned with latent-specific losses~\cite{radevski2023multimodal, chae2024kd, li2023decoupled}.
For example, L2 loss is used to align BEV feature space where teacher and student features share a geometric coordinate, while KL divergence loss is used for distributional latents~\cite{chae2024kd}.
In addition to adhering to such philosophy, we attempt to keep our framework generalizable.
In that sense, we have chosen the core distillation features by their general context: intra-agent modality fusion $Q$ and inter-agent interaction $H$.

Table~\ref{table:ablation_KD_components} studies the effects of each $\mathcal{L}_\text{KD}^\text{L}$ and $\mathcal{L}_\text{KD}^\text{G}$ on overall performance gain.
First, $\mathcal{L}_\text{KD}^\text{L}$ guides the student model to learn and encode the comprehensive knowledge of $Q_\mathcal{T}$ on agent's motion intent with its limited modality.
$\mathcal{L}_\text{KD}^\text{G}$ then further reinforces the student model by inducing the model to effectively encode inter-agent interactions on the heterogenous latents acquired from $Q_\mathcal{T}$.
For both HiVT and MART, $\mathcal{L}_\text{KD}^\text{L}$ is shown to be most important as it aligns the most fundamental latents that represent agents' motion intent.
Additional use of $\mathcal{L}_\text{KD}^\text{G}$ corroborates the distillation by guiding the model to learn incorporating interaction among these intents.
As a result, incorporating both $\mathcal{L}_\text{KD}^\text{L}$ and $\mathcal{L}_\text{KD}^\text{G}$ results in greatest improvement.
\\
\textbf{Does language assist in modeling human interactions?}\\
Human locomotion depends on two major factors: intra-agent motion intent and inter-agent interaction.
While recent works attempt to leverage agent-descriptive multi-modality in modeling motion intent~\cite{salzmann2023hst, saadatnejad2023social}, no works have explored utilizing explicit interaction cues other than global positional vectors.

To the best of our knowledge, we are the first to investigate the impact of incorporating explicit text as supplementary guidance in modeling agent interaction.
Table~\ref{table:ablation_interaction_text} demonstrates the improvements with the use of interaction text caption $\mathcal{S}_\text{R}$ on various models.
The most significant improvement is observed with the $\mathcal{X}$ model, which operates with least modalities. 
In this setting where data is scarce for inferring agent motion intent, the interaction text compensates for missing information by conveying implicit action intent cues embedded within the behavioral descriptions.
Other models also consistently improve with $\mathcal{S}_\text{R}$, with the performance gain decreasing as the number of modalities increases.
For KD, building a $\mathcal{X}+\mathcal{P}+\mathcal{S}$ model with the text-based guidance improved the teacher model by 0.9\%, followed by 0.51\% gain in its corresponding student.
Overall, we observe the broad impact of interaction text from building a competent teacher to distilling its knowledge.
\\
\textbf{What kind of language description is most helpful?}\\
From previous sections, we discovered that language plays the most important role in constructing a comprehensive teacher model.
Then, what kind of text caption helps the most?
To answer this question, we use VLM to generate captions based on diverse prompts shown in Tab.~\ref{table:prompts}.
We generate text captions for each visible agent in the text annotation-free SIT dataset.
Each prompt is selected to leverage different abilities of VLM.
Prompt 1 is the same format as JRDB annotation, describing the past behavior.
Prompt 2 incorporates the scene context by specifying any surrounding obstacle as text captions for ETH/UCY.
Prompt 3 takes a predictive stance on human motion by captioning the VLM's prediction on future human motion.

Evaluation on $\mathcal{X}+\mathcal{S}$ teacher model shows that incorporating the scene context yields the best performance.
Indeed, hints about the presence of an obstacle offer insight into the agent's future trajectory direction, whereas descriptive captions primarily focus on the agent's dynamic or static nature.
Comparing between prompt 1 and prompt 3, the $\mathcal{X}+\mathcal{S}$ teacher model shows comparable performance, showing that VLM's prediction of agent behavior accurately reflects the corresponding past motion.
With KD, however, knowledge from prompt 1 resulted in most superior performance due to the straightforward correlation between past pose and text describing the past motion.
For prompt 2, map context was unable to be inferred upon KD since there is no distinctive correlation between past pose and map context.
While KD with prompt 3 shows nearly comparable result compared to prompt 1, a direct correlation between pose and text as in prompt 1 yields greatest improvement upon KD between these latents.

\begin{table}[]
\caption{Ablation study on interaction text $\mathcal{S}_\text{R}$ using the HiVT model under various modality protocols. Experimented on JRDB. $\mathcal{X}$ column indicates the additional modalities used alongside $\mathcal{X}$.}
\vspace{-7pt}
\resizebox{\columnwidth}{!}{%
\begin{tabular}{|c|c|c|cccccc|c|}
\hline
$+\mathcal{S}_\text{R}$ & \begin{tabular}[c]{@{}c@{}}w/ \\ KD\end{tabular}  & $\mathcal{X}$ & ADE & \text{ADE}\textsubscript{2} & \text{ADE}\textsubscript{1} & FDE & \text{FDE}\textsubscript{2} & \text{FDE}\textsubscript{1} & \begin{tabular}[c]{@{}c@{}}Avg. \\ +\%\end{tabular} \\ \hline
 &  & \multirow{2}{*}{-} & 0.221 & 0.240 & 0.342 & \textbf{0.432} & 0.467 & 0.632 & \multirow{2}{*}{\textbf{+4.57}} \\
\checkmark &  &  & \textbf{0.218} & \textbf{0.228} & \textbf{0.316} & \textbf{0.432} & \textbf{0.445} & \textbf{0.576} &  \\ \hline
 &  & \multirow{2}{*}{$+\mathcal{P}$} & 0.229 & 0.242 & 0.364 & 0.441 & 0.465 & 0.659 & \multirow{2}{*}{+3.59} \\
\checkmark &  &  & \textbf{0.226} & \textbf{0.237} & \textbf{0.334} & \textbf{0.436} & \textbf{0.458} & \textbf{0.614} &  \\ \hline
 &  & \multirow{2}{*}{$+\mathcal{S}_\text{A}$} & 0.233 & 0.238 & 0.275 & 0.450 & 0.458 & 0.516 & \multirow{2}{*}{+0.98} \\
\checkmark &  &  & \textbf{0.224} & \textbf{0.230} & 0.285 & \textbf{0.432} & \textbf{0.444} & 0.540 &  \\ \hline
 &  & \multirow{2}{*}{\begin{tabular}[c]{@{}c@{}}$+\mathcal{P}$\\ $+\mathcal{S}_A$\end{tabular}} & 0.225 & 0.231 & \textbf{0.278} & 0.431 & 0.442 & \textbf{0.515} & \multirow{2}{*}{+0.90} \\
\checkmark &  &  & \textbf{0.220} & \textbf{0.227} & 0.280 & \textbf{0.423} & \textbf{0.435} & 0.521 &  \\ \hline
 & \checkmark & \multirow{2}{*}{$+\mathcal{P}$} & \textbf{0.227} & \textbf{0.238} & 0.317 & \textbf{0.444} & \textbf{0.461} & 0.580 & \multirow{2}{*}{+0.51} \\
\checkmark & \checkmark &  & 0.232 & 0.239 & \textbf{0.308} & 0.445 & 0.464 & \textbf{0.560} &  \\ \hline
\end{tabular}
}
\label{table:ablation_interaction_text}
\vspace{-3pt}
\end{table}

\begin{table}[]
\caption{Ablation on VLM prompts for acquiring text captions on SIT dataset. HiVT $\mathcal{X}+\mathcal{P}$ is distilled from $\mathcal{X}+\mathcal{S}$ for each.}
\vspace{-7pt}
\resizebox{\columnwidth}{!}{%
\begin{tabular}{|c|cccccc|}
\hline
\multicolumn{1}{|l|}{} & ADE & \text{ADE}\textsubscript{2} & \text{ADE}\textsubscript{1} & FDE & \text{FDE}\textsubscript{2} & \text{FDE}\textsubscript{1} \\ \hline
Prompt 1 & \multicolumn{6}{c|}{What is the person doing?} \\ \hline
$\mathcal{X}$ + $\mathcal{S}$ & \cellcolor[HTML]{DADADA}0.521 & \cellcolor[HTML]{DADADA}0.525 & \cellcolor[HTML]{DADADA}0.533 & \cellcolor[HTML]{DADADA}0.978 & \cellcolor[HTML]{DADADA}0.984 & \cellcolor[HTML]{DADADA}1.005 \\
$\mathcal{X}$ + $\mathcal{P}+$ KD & \cellcolor[HTML]{DADADA}\textbf{0.414} & \cellcolor[HTML]{DADADA}\textbf{0.444} & \cellcolor[HTML]{DADADA}\textbf{0.500} & \cellcolor[HTML]{DADADA}\textbf{0.789} & \cellcolor[HTML]{DADADA}\textbf{0.853} & \cellcolor[HTML]{DADADA}\textbf{0.951} \\ \hline
Prompt 2 & \multicolumn{6}{c|}{Is there any obstacle in front of the person?} \\ \hline
$\mathcal{X}$ + $\mathcal{S}$ & \textbf{0.506} & \textbf{0.508} & 0.531 & \textbf{0.950} & \textbf{0.959} & \textbf{0.991} \\
$\mathcal{X}$ + $\mathcal{P}+$ KD & 0.418 & 0.444 & \textbf{0.500} & 0.795 & 0.854 & 0.960 \\ \hline
Prompt 3 & \multicolumn{6}{c|}{What will the person do in the future?} \\ \hline
$\mathcal{X}$ + $\mathcal{S}$ & 0.520 & 0.522 & \textbf{0.530} & 0.989 & 0.990 & 1.003 \\
$\mathcal{X}$ + $\mathcal{P}+$ KD & \textbf{0.414} & 0.448 & 0.535 & 0.796 & 0.865 & 1.032 \\ \hline
\end{tabular}
}
\label{table:prompts}
\vspace{-9pt}
\end{table}

\section{Conclusion}

In this work, we present a multi-modal knowledge distillation framework for human trajectory forecasting. First, we demonstrate that incorporating additional modalities such as 3D pose and text significantly enhances forecasting performance.
However, as expensive modality such as text is not readily available in most applications, we additional perform KD on models with limited modalities.
Student models with only trajectory input or pose as sole additional modality is distilled upon a teacher trained with full range of modalities.
Student models consistently improve in both instantaneous and full observation metrics on all ego and BEV-view datasets, confirming the generalized competence of our KD framework.
\newline\noindent\textbf{Acknowledgements}
This work was supported by the National Research Foundation of Korea (NRF) grant funded by the Korea government (MSIT) (NRF2022R1A2B5B03002636).

{
    \small
    \bibliographystyle{ieeenat_fullname}
    \bibliography{main}
}


\end{document}